\documentclass[letterpaper, 10 pt, conference]{ieeeconf}  % Comment this line out if you need a4paper

\IEEEoverridecommandlockouts                              % This command is only needed if 
\usepackage{amsmath,amsfonts}
\usepackage{algorithm}
\usepackage{algpseudocode}
\usepackage{hyperref}
\hypersetup{hidelinks}
\usepackage{array}
\usepackage{courier}
\usepackage{subfigure}
\usepackage{textcomp,soul,color,xcolor}
\usepackage{stfloats}
\usepackage{url}
\usepackage{verbatim}
\usepackage{graphicx}
\usepackage{booktabs}
\usepackage{multirow}
\usepackage{cite}

\usepackage{bm}
\usepackage{xcolor}
\newcommand{\MYhref}[3][blue]{\href{#2}{\color{#1}{#3}}}%

\title{\LARGE \bf
LMMCoDrive: Cooperative Driving with Large Multimodal Model
}

\author{Haichao Liu, Ruoyu Yao, Zhenmin Huang, Shaojie Shen, and Jun Ma
\thanks{Haichao Liu and Ruoyu Yao are with the Robotics and Autonomous Systems Thrust, The Hong Kong University of Science and Technology (Guangzhou), China (e-mail: hliu369@connect.hkust-gz.edu.cn, ryao092@connect.hkust-gz.edu.cn).}
\thanks{Zhenmin Huang, Shaojie Shen and Jun Ma are with the Department of Electronic and Computer Engineering, The Hong Kong University of Science and Technology, Hong Kong SAR, China (e-mail: zhuangdf@connect.ust.hk; eeshaojie@ust.hk; jun.ma@ust.hk).}
}

\begin{document}

\maketitle
\thispagestyle{empty}
\pagestyle{empty}

%%%%%%%%%%%%%%%%%%%%%%%%%%%%%%%%%%%%%%%%%%%%%%%%%%%%%%%%%%%%%%%%%%%%%%%%%%%%%%%%
\begin{abstract}
% Addressing the intricate challenges of Decentralized Cooperative Scheduling and Motion Planning in Autonomous Mobility-on-Demand (AMoD) systems, this paper presents LMMCoDrive, a novel approach that harnesses the capabilities of a Large Multimodal Model (LMM) to augment the traffic efficiency of such systems within dynamic urban environments. Our work introduces a comprehensive LMM-enhanced framework designed to seamlessly integrate scheduling and motion planning processes, thereby ensuring the effective operation of Cooperative Autonomous Vehicles (CAVs). The spatial relationship between CAVs and passenger requests is abstracted into a bird's-eye view (BEV) in order to fully exploit the potential of the LMM. Besides, by adopting a Model Predictive Control (MPC) strategy, we cautiously refine the trajectories for each CAV, incorporating safety constraints within the optimization problems to adeptly manage collision avoidance. Further, we apply a decentralized optimization strategy empowered by the Alternating Direction Method of Multipliers (ADMM) into AMoD, which is all under the aegis of LMM for the graph evolution of the CAVs. Through rigorous simulation, our results underscore the pivotal role of LMM in optimizing CAV scheduling and enhancing the graph evolution for the decentralized cooperative optimization process of each vehicle. This marks a substantial stride towards achieving practical, efficient, and safe AMoD systems, poised to revolutionize urban transportation. 
To address the intricate challenges of decentralized cooperative scheduling and motion planning in Autonomous Mobility-on-Demand (AMoD) systems, this paper introduces LMMCoDrive, a novel cooperative driving framework that leverages a Large Multimodal Model (LMM) to enhance traffic efficiency in dynamic urban environments. This  framework seamlessly integrates scheduling and motion planning processes to ensure the effective operation of Cooperative Autonomous Vehicles (CAVs). The spatial relationship between CAVs and passenger requests is abstracted into a Bird's-Eye View (BEV) to fully exploit the potential of the LMM. Besides, 
% by adopting a Model Predictive Control (MPC) strategy, 
trajectories are cautiously refined for each CAV while ensuring collision avoidance through safety constraints. A decentralized optimization strategy, facilitated by the Alternating Direction Method of Multipliers (ADMM) within the LMM framework, is proposed to drive the graph evolution of CAVs. Simulation results demonstrate the pivotal role and significant impact of LMM in optimizing CAV scheduling and enhancing decentralized cooperative optimization process for each vehicle. This marks a substantial stride towards achieving practical, efficient, and safe AMoD systems that are poised to revolutionize urban transportation.
The code is available at \MYhref[black]{https://github.com/henryhcliu/LMMCoDrive}{https://github.com/henryhcliu/LMMCoDrive}.
\end{abstract}

%%%%%%%%%%%%%%%%%%%%%%%%%%%%%%%%%%%%%%%%%%%%%%%%%%%%%%%%%%%%%%%%%%%%%%%%%%%%%%%%
\section{Introduction}

Autonomous Mobility-on-Demand (AMoD) Systems represent a transformative approach to urban transportation, which aim to efficiently and safely navigate complex urban environments to fulfill passenger requests with fleets of Cooperative Autonomous Vehicles (CAVs)~\cite{horl2019dynamic}. These systems promise to revolutionize urban landscapes by mitigating traffic congestion, enhancing accessibility, and reducing transportation-related emissions. However, orchestrating such systems presents significant challenges, particularly in the realm of task scheduling and cooperative motion planning~\cite{ding2023mechanism}. The inherent dynamic nature of AMoD systems necessitates continuous path adjustments for CAVs to manage both assigned and future tasks, thereby complicating vehicle scheduling in response to passenger requests within the intricate configurations of urban traffic systems~\cite{aalipour2024modeling}.

Historically, research has been carried out to segregate the focus on scheduling, particularly on how CAVs are dispatched to meet user demands. Yet, the critical need for the integration of scheduling and motion planning is neglected. This separation overlooks the essential requirement for simultaneous motion planning and mutual avoidance among assigned vehicles, which is crucial for the practical deployment of AMoD systems in complex urban environments. This research gap signifies a pressing need for possible integrated solutions.

The challenges of integrating intelligent scheduling with real-time cooperative motion planning for AMoD systems are threefold. First and foremost, there is the formidable challenge of developing a universally applicable scheduling strategy that can effectively navigate the intricacies of urban traffic systems, particular for rule-based and  learning-based methods~\cite{javanshour2021performance}. Secondly, ensuring an agile response from the motion planning level to changes at the scheduling level is essential for the dynamic urban traffic environment~\cite{veres2019deep}. Lastly, achieving fast cooperative motion planning with a large number of CAVs, without compromising on safety or efficiency, introduces another level of complexity~\cite{karlsson2018multi}.
\begin{figure}[t]
    \centering
    \includegraphics[width=\linewidth]{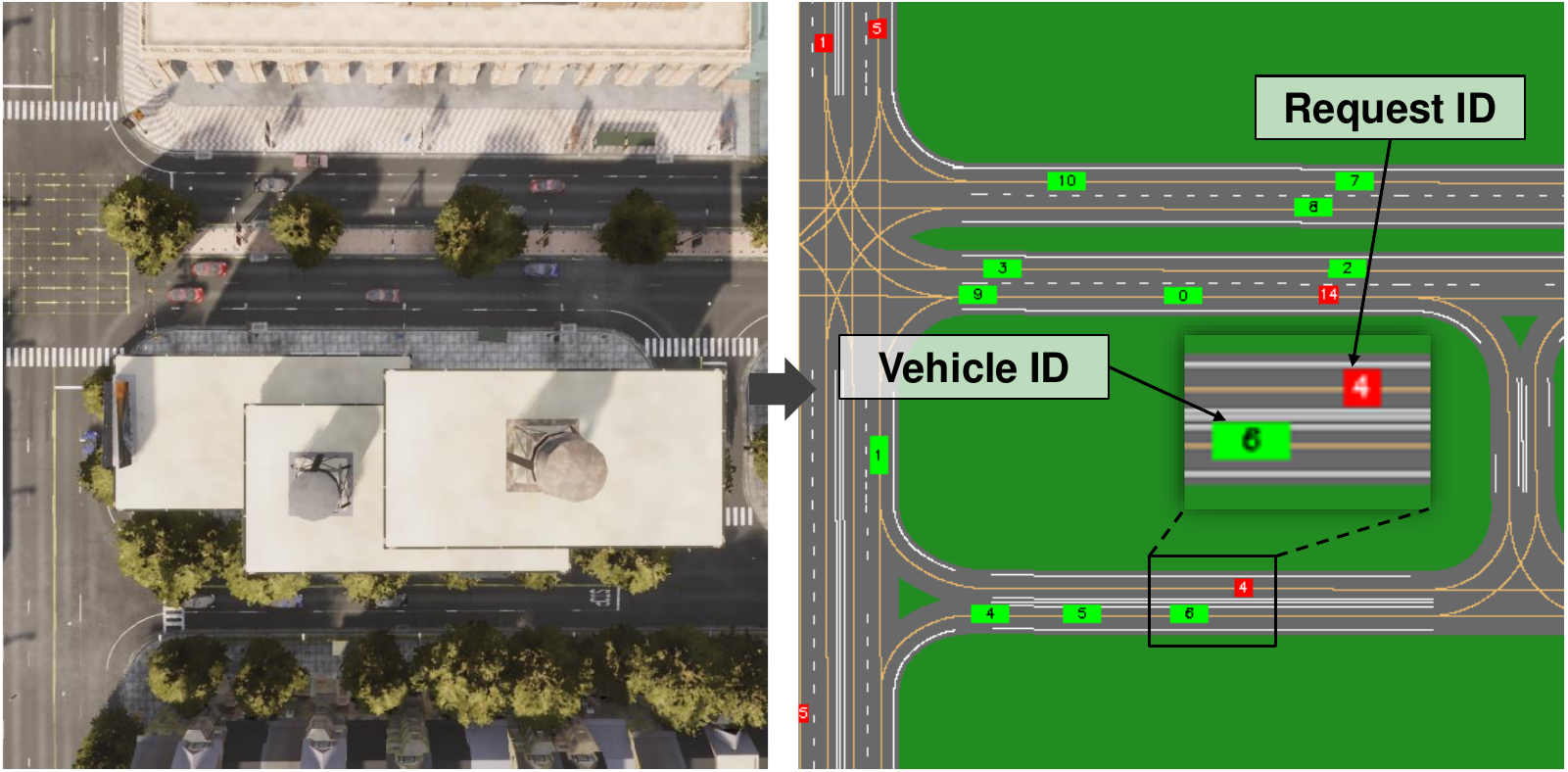}
    \caption{Demonstration of the autonomous mobility-on-demand system in an urban scenario. The red squares denote the requests from passengers, while the free vehicles are supposed to be scheduled by the LMM. The abstracted bird-eye-view graphics are generated along with the supplementary textual information to be sent to the LMM for multiple decisions.}
    \label{fig:demo_bev}
\end{figure}
In light of these challenges, this work introduces {LMMCoDrive}, a novel approach leveraging a Large Multimodal Model (LMM) to enhance decentralized cooperative scheduling and motion planning of AMoD systems. Our primary contributions are as follows:

\begin{itemize}
    \item We propose a novel LMM-enhanced framework that integrates scheduling and motion planning processes for CAVs, in which we leverage an informative BEV of the AMoD system to exploit the potential of LMMs in cooperative driving tasks.
    % which ensures safe and efficient operations of the CAVs in dynamic urban environments.
    % \item We refine the path for each CAV with the MPC, incorporating hard constraints on the optimization problems and enhancing the agile scheduling capability.
    \item We present a decentralized optimization algorithm via the Alternating Direction Method of Multipliers (ADMM) under the guidance of LMM for graph evolution. It efficiently splits the CAVs into several sub-graphs, leading to smaller-sized OCPs for high computational efficiency.
    \item Experimental results demonstrate the effectiveness of LMMCoDrive in optimizing the scheduling of CAVs and also in the graph evolution process for the cooperative driving task. The results attained underscore a significant advancement for LMMs towards the realization of practical AMoD systems.
\end{itemize}

% which is the first of its kind to our best knowledge. 
% Thus, our work is beneficial to the autonomous driving community.
\section{Related Works}

% This section delves into existing methodologies and highlights our contributions in the domains of scheduling and cooperative motion planning for Autonomous Mobility-on-Demand (AMoD) systems, underscoring the novelty of our Large Multimodal Model (LMM)-enhanced approach.

\subsection{Scheduling Methods for AMoD}

Recent studies have delved into the application of Reinforcement Learning (RL) in the realm of scheduling for AMoD systems. For example, the integration of multi-agent Soft Actor-Critic with weighted bipartite matching achieves small-scale scheduling of the agents~\cite{enders2023hybrid}. Despite their innovative strategies, RL-based methods grapple with inherent drawbacks. The complexity of AMoD scheduling, which encompasses numerous variables and scenarios, makes it challenging to define a clear and effective reward function~\cite{he2023robust}. Furthermore, the scarcity of real-world datasets for task scheduling of AMoD systems limits the practical applicability and training of RL models.
To navigate the task allocation process, various rule-based methods have been proposed, such as Distance First (DF), Idle First (IF), and Priority First (PF)~\cite{dandl2021regulating,guo2022rebalancing}. These methods, guided by predefined rules, offer simplicity and interpretability but may lack the flexibility and scalability needed to address the inherent dynamic nature of urban transportation systems.

The advent of Large Language Model (LLM) has paved the way for advancements in task scheduling, owing to their prowess in open-world comprehension, reasoning, and few-shot acquisition of common sense knowledge~\cite{wei2022chain,huang2023voxposer, zu2024language, yao2024tree}. Notable efforts like DiLu leverage textual descriptions of the environment for single-vehicle decision-making~\cite{wen2023dilu}. The potential of LLM to make contextual inferences and decisions heralds a promising direction for scheduling of AMoD systems. Nonetheless, the application of multimodal LLM or LMM in cooperative driving for AMoD systems has yet to be extensively explored~\cite{cui2024survey}. With their information representation capabilities, especially in dealing with images, LMMs have the potential to greatly improve the depiction and comprehension of urban transportation systems, which could underscore a crucial research gap.

\subsection{Cooperative Motion Planning Methods for AMoD}

After the scheduling process of the AMoD system, each CAV is required to plan a feasible trajectory to be followed. Optimization-based approaches to cooperative motion planning typically employ a formulation as an Optimal Control Problem (OCP). This methodological framework offers a precise mathematical representation, strong interpretability, and guarantee of optimality~\cite{li202optimization, huang2024parallel}. For CAVs with nonlinear vehicle models, OCPs are typically solved using established nonlinear programming solvers, such as interior point optimizer (IPOPT) and Sequential Quadratic Programming (SQP). Moreover, the iterative linear quadratic regulator (iLQR) method, benefiting from Differential Dynamic Programming (DDP), addresses nonlinear optimization problems by retaining only the first-order term of dynamics through Gauss-Newton approximation~\cite{lee2022gpu}. However, traditional iLQR struggles to directly manage inequality constraints related to collision avoidance and physical limitations in this situation. To address this challenge, advanced variations such as control-limited DDP and constrained iLQR have been suitably developed~\cite{Ma2022Alternating,chen2019autonomous,ma2023local}.

To alleviate the computational burdens of cooperative driving in large-scale AMoD systems, ADMM emerges as a potent solution by decomposing the principal optimization problem into several sub-problems. Its parallel and distributed nature renders ADMM particularly suitable for cooperative motion planning of CAVs.  For instance, dual consensus ADMM has been deployed to facilitate fully parallel optimization framework for cooperative motion planning of CAVs~\cite{huang2023decentralized}. This approach significantly distributes computational efforts across all entities and attains real-time performance. Despite these advancements, existing methods fall short of supporting large-scale cooperative driving due to the fully connected nature of agents. Our prior research, which leverages the sparsity attribute of optimization problems for involved vehicles~\cite{liu2024improved}, indicates a promising pathway to enhance computational efficiency for AMoD systems, and this highlights another crucial research gap to be addressed.

\section{Problem Formulation}
We consider a discrete-time system with $N$ single-occupancy vehicles, denoted by $\mathcal{V}$. Customers randomly arrive and await transportation to their destinations. Unlike traditional first-come-first-serve (FCFS) methods, our system determines the order of service, prioritizing efficiency and directionality akin to ride-sharing practices. 
% This approach, suitable for systems without fixed "stations," utilizes mobile app requests for customer pickups anywhere along the roadside, which is identical to real-world customer-requiring situations.

The representation of CAVs' relationships using graph theory, where nodes and edges signify vehicles and communication links respectively, facilitates a detailed analysis of their interactions. We define an undirected graph $\mathcal{G} = (\mathcal{V}, \mathcal{E})$, with $\mathcal{V}$ as the set of vehicles and $\mathcal{E}$ as the communication links. Each vehicle $n^i \in \mathcal{V}$ has a state vector $\bm{z}^i$ and a control input vector $\bm{u}^i$. Subgraphs $\mathcal{H}$, representing vehicle interactions within specific communication ranges, evolve over cooperative driving planning horizons. 
% The total number of CAVs, $N$, and the distance matrix $\bm{D} = [a_{ij}]$ for $\mathcal{G}$ are introduced, with $N_k$ denoting the number of CAVs in each subgraph $\mathcal{H}_k$. 
An edge $(n^i, n^j) = d^{i,j} \in \mathcal{E}_h \subseteq \mathcal{E}$ exists if the inter-vehicle distance $d^{i,j}$ is less than the communication range $r^i_\text{tele}$. The degree of a node $n^i$, i.e., $|{n^i}| = \text{deg}(n^i)$, indicates its connectivity with the nodes in the set
\begin{equation}\label{eq:neighbor_node_set}
    \mathcal{N}^i = \{j\in\mathcal{N}\, | \,(i,j)\in \mathcal{E}_h\}.
\end{equation}
This graph-theoretic model supports the development of scheduling and cooperative motion planning algorithms for AMoD systems.
\begin{figure*}
    \centering
    \includegraphics[width=1\linewidth]{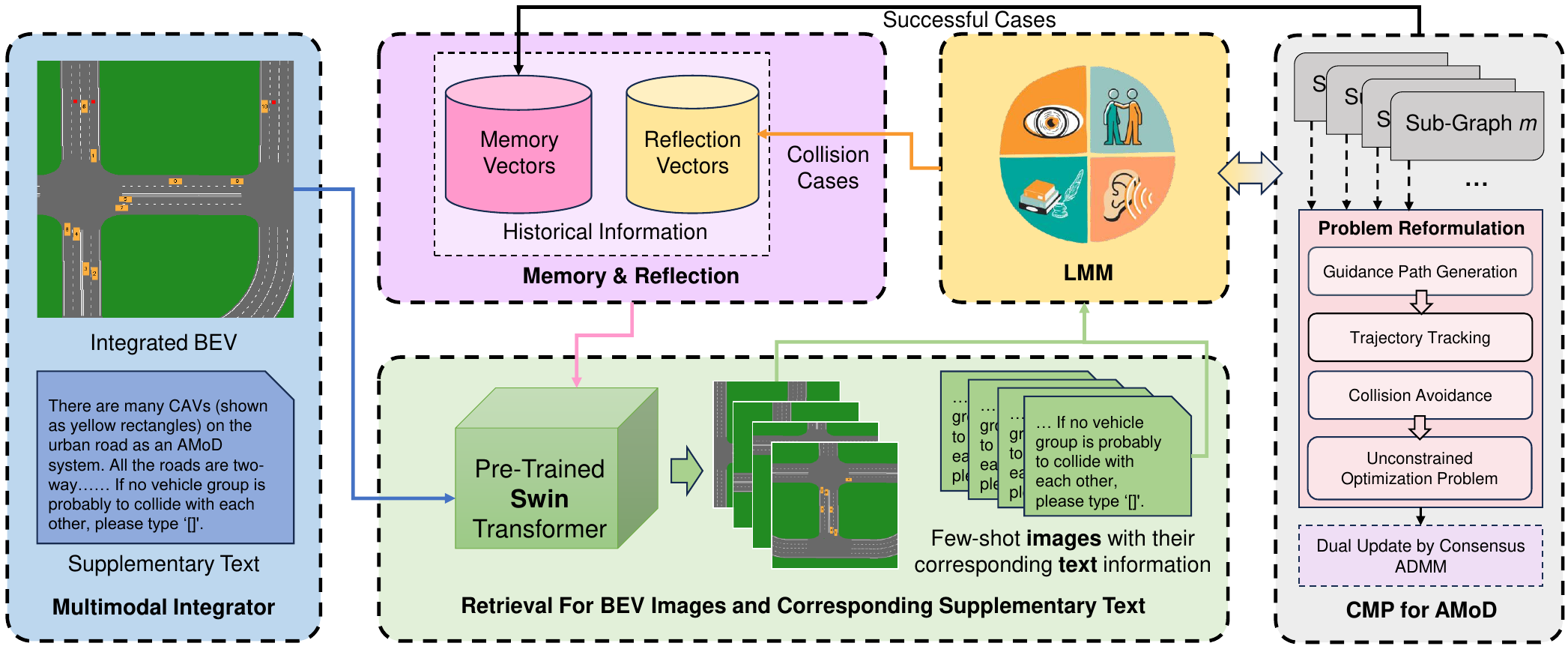}
    \caption{Overall architecture of LMMCoDrive. It is composed of a multimodal integrator, a memory and reflection module, a retrieval module for similar memory and reflections, and a decentralized cooperative driving module. CMP means cooperative motion planning for the CAVs in the AMoD system.}
    \label{fig:LMMCoDriveFramework}
\end{figure*}

\section{Method}

Our proposed LMMCoDrive scheme mainly contains two parts: First, an LMM-based scheduling agent, is responsible for the multimodal information gathering, memory, reflection, and retrieval of similar historical situations. Second, an LMM-guided cooperative driving module, provides low-level motion planning for the vehicles and ensures the safety.
The executing process is exhibited in Fig.~\ref{fig:LMMCoDriveFramework}.
% \subsection{LMM-based Scheduling}\label{subsec:LMM-based scheduling}
% \subsubsection{Multimodal Integrator}
% \subsubsection{Memory and Reflection Module}
% \subsubsection{Retrieval Module}
\subsection{LMM-Based Scheduling}

The proposed LMM-based scheduling framework is designed to integrate and process multimodal information for effective task assignment of AMoD systems. This section delves into Information Integrator, Memory and Reflection Module, and Retrieval Module successively.

\subsubsection{Multimodal Integrator}

The multimodal integrator is pivotal in synthesizing data from various sources to generate a comprehensive understanding of the urban traffic environment and AMoD system status. This module comprises two primary components: BEV image generation and supplementary text generation.

% At the core of this framework is the Multimodal Integrator, which plays a crucial role in generating a detailed representation of the traffic scene.
This module starts by creating BEV images that encapsulate critical elements of the urban traffic landscape. These elements include road boundaries, lanes, CAVs each tagged with a unique ID, the centerlines of road lanes, and pending requests from passengers marked with unique IDs. The creation of BEV images is instrumental in providing a visual foundation for the scheduling system's decision-making processes.

Following the generation of BEV images, the module undertakes the production of supplementary text. This text plays multiple roles, serving first as a SystemMessage that offers a deeper dive into the driving environment. It differentiates types of lane markings, delineates the driving layout, and categorizes vehicles into left-hand drive (LHD) and right-hand drive (RHD) based on their operational side and the side of the road they drive on. These objects are identified by exclusive IDs, enabling the accuracy of textual expression for the LMM.

\subsubsection{Memory and Reflection Module}

As shown in the middle of Fig.~\ref{fig:LMMCoDriveFramework}, this module consists of two parts: the memory vectors and the reflection vectors.
The memory module acts as a repository, storing historical images along with their corresponding textual information (i.e. prompting text and LMM response) in a stack. This stack is limited in size, ensuring that only the most recent and relevant information is retained for future reference.

Upon encountering collision cases, the reflection module feeds this multimodal data into the LMM for further analysis. The outcome of this reflection, alongside the original multimodal data, is stored back into the Memory module. This process enables the system to learn from past incidents and refine its decision-making capabilities over time.

\subsubsection{Retrieval Module}

% The primary function of the Retrieval module is to access the Memory module for similar past situations that can aid in few-shot inference for current scheduling decisions. By leveraging historical data, the system can make informed, context-aware decisions, improving both efficiency and safety in the AMoD system's operation. This module ensures that the LMM-based scheduling framework can adaptively respond to a wide range of scenarios by drawing on past experiences, enhancing its ability to manage the complex dynamics of urban traffic and AMoD system demands. 
The primary function of the retrieval module is to access the memory module for historical situations similar to the current context, thereby facilitating few-shot inference for scheduling decisions. By leveraging historical data, the system is capable of making informed, context-aware decisions that enhance both the efficiency and safety of the AMoD system's operations. This module ensures that the LMM-based scheduling framework can adaptively respond to a diverse array of scenarios by drawing upon past experiences, thereby improving its management of the complex dynamics of urban traffic and AMoD system demands.
% [to be expanded]
\begin{figure}
    \centering
    \includegraphics[width=1\linewidth]{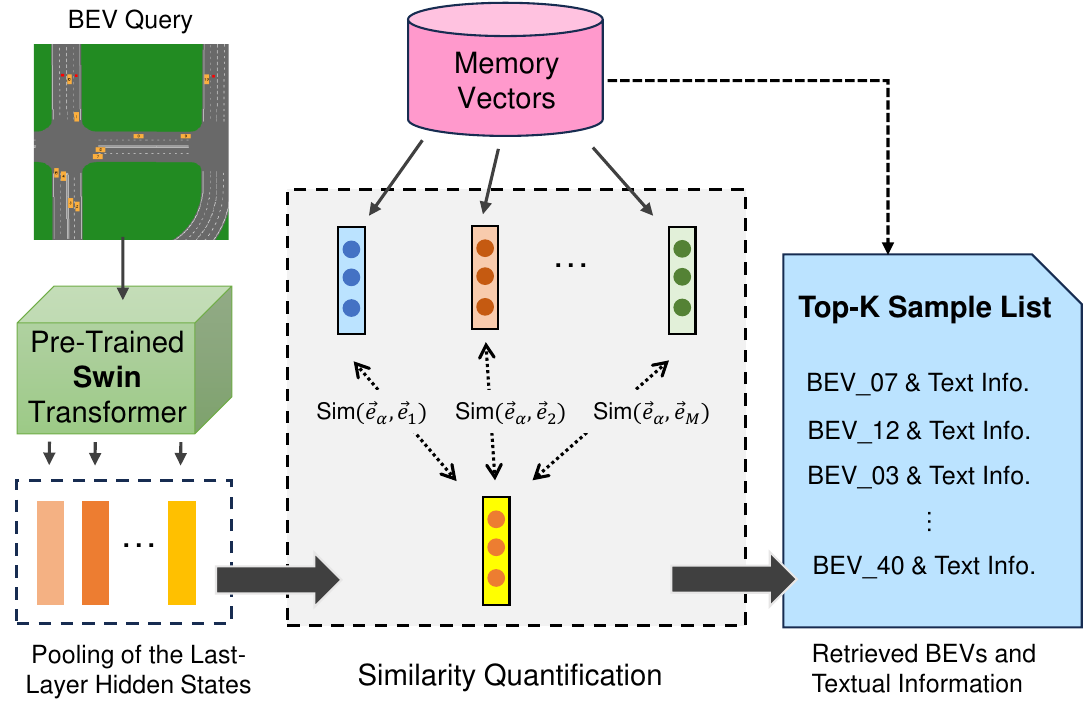}
    \caption{Illustration of the retrieval for LMMCoDrive. A BEV image for similarity query is embedded with the pre-trained swin-transformer, before the pair-wise similarity quantification with the memory vectors. The memory samples of Top-K similarities are retrieved to support the online reasoning.}
    \label{fig:swin retrieval}
\end{figure}

% To enable the retrieval process, we assign embeddings for each BEV image in the Memory module through the utilization of a pre-trained vision model. Swin-transformer~\cite{liu2021swin}, pre-trained on ImageNet-21k~\cite{deng2009imagenet}, is adopted for this study due to its superior capability in modeling multi-scale vision entities, which can properly extract features regarding both the macroscopic traffic scenario (intersection, multi-lane roadway, etc.) and microscopic vehicle status (vehicle IDs, position, orientation, etc.). The embedding is obtained by post-processing the last-layer hidden state of the classification token for summarizing the global feature representation. As shown in Fig. \ref{fig:swin retrieval}, for the online retrieval, the real-time BEV image is first embedded with the vision model, before evaluating its similarity with the memory samples using the cosine similarity score:
To facilitate the retrieval process, each BEV image in the Memory module is assigned an embedding using a pre-trained vision model. In this study, we adopt the Swin-Transformer~\cite{liu2021swin}, pre-trained on ImageNet-21k~\cite{deng2009imagenet}, owing to its superior capability in modeling multi-scale visual entities. This allows for the effective extraction of features pertinent to both macroscopic traffic scenarios (such as intersections and multi-lane roadways) and microscopic object statuses (including vehicle IDs, positions, and orientations). The embedding is derived by average pooling of the final-layer hidden states to summarize the global feature representation. As illustrated in Fig.~\ref{fig:swin retrieval}, during online retrieval, the real-time BEV image is first embedded using the vision model, and then its similarity to the memory samples is evaluated using the cosine similarity score:
\begin{equation}
    \text{Sim}(\vec{e}_\alpha, \vec{e}_\beta) = \frac{\vec{e}_\alpha \cdot \vec{e}_\beta}{\|\vec{e}_\alpha\| \|\vec{e}_\beta\|}, \forall \beta \in \{1,2,...,M\},
\end{equation}
% where $\vec{e}_\alpha$ denotes the embedding of the real-time BEV image for similarity query and $\vec{e}_\beta$ represents the embedding of any BEV image in the memory module, with $M$ standing for the memory size. After that, the BEV images of Top-K similarities are retrieved, which constitute few-shot examples along with the corresponding textual information to guide the online reasoning.
where $\vec{e}_\alpha$ denotes the embedding of the newly generated BEV image used for the similarity query, while $\vec{e}_\beta$ represents the embedding of any BEV image stored in the Memory module, and $M$ indicates the size of the memory. Following this, the BEV images with the Top-$K$ similarities are retrieved. This multimodal information forms the few-shot data that guide the online reasoning process.

\subsection{LMM-Guided Cooperative Driving}
% For the scheduled vehicle fleet of the AMoD system, a cooperative motion planning task is considered in one subgraph $\mathcal{H}$ generated by the LMM-guided graph evolution module, where $N$ CAVs drive on the urban road with their destinations and global path. We formulate the cooperative driving problem as an OCP:
For the scheduled AMoD vehicle fleet, we consider a cooperative motion planning task within subgraph $\mathcal{H}$ generated by the LMM-guided graph evolution module. We formulate this task, involving $N$ CAVs on urban roads with specified destinations and paths, as an OCP:
\begin{equation}
\label{NMPCOptProb}
\begin{array}{ll}
\underset{{\bm{z}_\tau^i}, {\bm{u}_\tau^i}}{\min} & \sum_{i=1}^N Q_i(\bm Z^i, \bm U^i)\\
\text { s.t.} & \boldsymbol z^i_{\tau+1}=f(\boldsymbol z^i_\tau,\boldsymbol u^i_\tau),  \\
& \bm z^i_{\tau+1}\in \mathcal{S}^i_\tau,\\
& -\boldsymbol {\underline u}^i \preceq \boldsymbol {u}_\tau^i \preceq \boldsymbol {\overline u}^i,\\
% & -\boldsymbol {\underline z}^i \preceq \boldsymbol z_\tau^i \preceq \boldsymbol {\overline z}^i,\\
% & -\boldsymbol {\underline z}^i \preceq \boldsymbol z_T^i \preceq \boldsymbol {\overline z}^i,\\
&\forall \tau \in \mathcal{T},\ \forall i \in \mathcal{N},
\end{array}
\end{equation}
% where $\mathcal{N}$ represents the set of index of CAVs in the traffic system of a subgraph $\mathcal H$. $\mathcal{T}=\{0,1,..., T-1\}$ is the temporal horizon within a cooperative motion planning episode. 
% In addition, $\mathcal{S}_\tau^i$ is the collision-free and bounded region of the $i$th CAV at time step $\tau$. ${\bm {\underline u}}^i$ and ${\bm {\overline u}}^i$ mean the lower and upper bound of the control input for $i$th CAV, respectively.
% Moreover, the objective of this OCP is to follow the reference path from the upstreaming scheduling assignment. For an individual vehicle, it can be expressed as
where $\mathcal{N}$ represents the set of indices of CAVs within a subgraph $\mathcal{H}$, and $\mathcal{T} = \{0, 1, \ldots, T-1\}$ denotes the temporal horizon for a cooperative motion planning episode. $\mathcal{S}_\tau^i$ defines the collision-free and bounded operational region for the $i$th CAV at time step $\tau$. ${\bm{\underline{u}}}^i$ and ${\bm{\overline{u}}}^i$ represent the lower and upper bounds of the control input for the $i$th CAV, respectively. The objective of this OCP is to enable the CAVs to follow their reference paths assigned by the upstream scheduling, which can be expressed as:

\begin{equation}
    Q_i(\bm z^i_\tau, \bm u^i_\tau) = \sum _{\tau=0}^T \| \bm z^i_\tau-\bm z^i_{\text{ref},\tau}\|^2_{\bm Q}+ \sum_{\tau=0}^
{T-1}\| \bm u^i_\tau\|^2_{\bm R},
\end{equation}
where $\bm Q$ and $\bm R$ are the weighting matrices balancing the path tracking and energy saving. 
% The expected outcome of the optimization problem (\ref{NMPCOptProb}) is to cooperatively follow the reference states with different weights while complying with the vehicle model and staying in the safe region $\mathcal{S}^i_\tau$ to avoid collisions with each other.
\begin{algorithm}[t]
\caption{Decentralized Optimization via Consensus ADMM in One Subgraph $\mathcal H(\mathcal{N},\mathcal{E}_h)$}\label{alg:alg2}
\begin{algorithmic}[1]
\State \textbf{initialize} $\{x^i_\tau,u^i_\tau\}^T_{\tau=0},\{p^{i,0},y^{i,0},z^{i,0},s^{i,0}\}, \forall i\in \mathcal{N}$
\State \textbf{choose} $\sigma,\rho >0$
\State \textbf{repeat}:
\State \hspace{0.3cm} Send $\{z^i_\tau\}_{\tau=1}^T$, receive $\{z^j_\tau\}_{\tau=1}^T$ from $j\in\mathcal{N}^i$
\State \hspace{0.3cm} Compute $k^i$, $J^i$, $\{A^i_\tau\}^{T-1}_{\tau=0}$, $\{B^i_\tau\}^{T-1}_{\tau=0}$
\State \hspace{0.3cm} \textbf{reset} $p^{i,0}=s^{i,0}=0, y^{i,0}=y^\text{last},x^{i,0}=x^\text{last}$
\State \hspace{0.3cm} \textbf{reset} $y^{i,0}_{[2]} = x^{i,0}_{[2]}=0$
\State \hspace{0.3cm} \textbf{repeat}: for all $i \in \mathcal{N}$
\State \hspace{0.6cm} Send $y^{i,k}$, receive $y^{j,k}$ from $j\in\mathcal{N}^i$
% \State \hspace{0.6cm} Steps 4-7 of Algorithm~\ref{alg:alg1} for $[\, \cdot \,]^{i,k}_{[1]}$
\State \hspace{0.6cm} Broadcast $y^{i,k}$ to the vehicles in $\mathcal{N}^i$
\State \hspace{0.6cm} $p^{i,k+1}=p^{i,k}+\rho\sum_{j\in \mathcal{N}^i}(y^{i,k}-y^{j,k})$
\State \hspace{0.6cm} $s^{i,k+1}=s^{i,k}+\sigma(y^{i,k}-x^{i,k})$
\State \hspace{0.6cm} $r^{i, k+1} = \sigma x^{i, k}+\rho \sum_{j\in \mathcal{N}^i}\left(y^{i, k}+y^{j, k}\right)$
\Statex \hspace{1.6cm} $-(k^i+p^{i, k+1}+s^{i, k+1})$
\State \hspace{0.6cm} Perform (\ref{subeq:imp1})-(\ref{subeq:imp3}) for $ \alpha^{i,k}_{[2,i]}, \alpha\in\{ p, s,  r\}$
\State \hspace{0.6cm} Perform (\ref{eq:sub1})-(\ref{eq:sub3}) for $ \alpha^{i,k}_{[2,j]},\alpha\in\{ p, s,  r\},j\in\mathcal{N}^i$
\State \hspace{0.6cm} Compute $z^{i, k+1}$ by solving LQR problem (\ref{eq:quadOptProb})
\State \hspace{0.6cm} $y^{i, k+1}_{[1]} = 2\gamma\left(\hat J^i z^{i, k+1}_{[1]}+r^{i, k+1}_{[1]}\right)$
\State \hspace{0.6cm} ${x}_{[1]}^{i, k+1} = \Pi_{\mathcal{R}^{\circ}_+}\left(\frac{1}{\sigma}s^{i, k+1}_{[1]}+ y^{i, k+1}_{[1]}\right)$
\State \hspace{0.6cm} Perform (\ref{subeq:imp4}-\ref{subeq:imp5}) for $ \alpha^{i,k}_{[2,i]}, \alpha\in\{ y, x\}$
\State \hspace{0.6cm} Perform (\ref{eq:sub4}-\ref{eq:sub5}) for $\alpha^{i,k}_{[2,j]}, \alpha\in\{ y,x\},j\in\mathcal{N}^i$
\State \hspace{0.6cm} $k=k+1$
\State \hspace{0.3cm} \textbf{until} number of iteration steps exceeds $k_\text{max}$
\State \hspace{0.3cm} Update $\{z^i_\tau,u^i_\tau\}^T_{\tau=0}$
\State \textbf{until} termination criterion is satisfied
\end{algorithmic}
\end{algorithm}

\subsubsection{LMM-Guided Graph Evolution}
% The graph evolution process is crucial to simplify the individual OCPs for the motion planning of the fleet of CAVs. We use a similar policy with the scheduling process for the AMoD. Note that the multimodal information integration module is slightly different from the equivalent module for scheduling. As the purpose of the graph evolution is to generate the groups of CAVs that have collision risks, the passengers' requests should not appear in the generated BEV, and the decision of the LMM should also be changed accordingly, from the request and response pair to the vehicle groups having collision risk.
% We generate the task description along with the retrieved similar multimodal conversations as a few-shot description. After postprocessing of the LMM's answer, the vehicle groups having collision risk are generated by the graph evolution guided by the LLM.
The graph evolution is essential for simplifying the OCPs involved in the motion planning of the CAV fleet. We employ a policy similar to that used in the scheduling process. However, the multimodal information integrator differs slightly from its scheduling counterpart. Since the graph evolution aims to identify groups of CAVs with potential collision risks, passenger requests are excluded from the BEV images, and the LMM's decision-making shifts from handling request-response pairs to identifying vehicle groups at risk of collision.

We generate a task description alongside retrieved similar multimodal contexts to form a few-shot example. Following the postprocessing of the LMM's output, the graph evolution, guided by the LLM, produces the groups of vehicles with collision risks. The corresponding adjacency matrix records the above information effectively, as depicted in Algorithm~\ref{alg:alg3}.
\subsubsection{ADMM-Based Cooperative Motion Planning}
% We treat the cooperative driving for the CAVs after scheduling and graph evolution as a constraint combinatorial optimization problem. 
In order to reduce the size of the OCP and provide an agile response, MPC is used with a planning horizon $T_p=15$, and an executing horizon $T_e=10$. The dual formulation of the OCP derived from (\ref{NMPCOptProb}) is as follows:
\begin{equation}
    \begin{aligned}
        \min_{\bm {\Delta Z}^1,...,\bm{\Delta Z}^N} &\sum_{i=1}^N F^i(\bm {\Delta Z}^i)+\mathcal{I}_{\mathcal{K}}\left(\bm h\right)\\
        \text{s.t. } & \sum_{i=1}^N \left(\bm J^i\bm {\Delta Z}^i-\bm k^i\right)=\bm h,
    \end{aligned}
\label{ConstrainedNewProb}
\end{equation}
where $\bm{\Delta Z}^i$ is the concatenated vector of the state and control variables sequentially. ${F}^i$ is the host cost for vehicle $i$, $\mathcal{I}_\mathcal{K}$ is the indicator function for the mutual avoidance and bounded constraints. More detailed information can be referred to in~\cite{grontas2022distributed}.
% Based on the dual consensus ADMM algorithm proposed in~\cite{grontas2022distributed}, leveraging the Lagrangian of (\ref{ConstrainedNewProb}), the cooperative motion planning of the vehicles in the AMoD system is derived as shown in Algorithm~\ref{alg:alg2}. It contains two main processes, \textit{dual update} for the dual variables using steps 10-15 and 17-20, and \textit{primal update} to explore the optimal states and control inputs assuming the variables of other vehicles remain constant using step 16 in Algorithm~\ref{alg:alg2}.
Leveraging the Lagrangian of (\ref{ConstrainedNewProb}), the problem is derived with the dual variables ${\bm p, \bm s, \bm r,\bm y,\bm x}$. \textit{Dual update} is achieved by lines~10-15 and 17-20, while \textit{primal update} is realized using line~16, as shown in Algorithm~\ref{alg:alg2}.
% It contains two main processes, \textit{dual update} for the dual variables using steps 10-15 and 17-20, and \textit{primal update} to explore the optimal states and control inputs assuming the variables of other vehicles remain constant using step 16 in Algorithm~\ref{alg:alg2}.

\begin{algorithm}[t] \caption{Cooperative Scheduling and Motion Planning for CAVs with LMM}\label{alg:alg3}
\begin{algorithmic}[1] 
\State \textbf{schedule} target positions of all CAVs in $\mathcal{G}$ using LMM
\State \textbf{determine} optimal route using A* for each CAV
\State \textbf{smooth} the waypoints in each route by the Savitzky-Golay filter
\While{$\exists$ requests in the AMoD}: 
\State Execute LMM-guided graph evolution to distribute the CAVs to $\mathcal{H}^k$
\For{subgraph $\mathcal{H}^i$ in subgraphs list} 
\State \textbf{search} nearest waypoints for each CAV at each time step $\tau\in \{0,1,...,T_s\}$ using KD-tree
\State Execute Algorithm \ref{alg:alg2} to obtain CAVs' trajectories
\EndFor
\State \textbf{perform} first $T_e$ steps of the CAVs in $\mathcal{G}$
\State \textbf{feedback} at time step $T_e$
\State \textbf{relay} the states of all the CAVs in $\mathcal{G}$
\EndWhile
\end{algorithmic}
\end{algorithm}
To enhance computational efficiency in the dual update process, we define two distinct sets of variables: one set associated with the target vehicle and another set pertaining to the surrounding vehicles that pose a collision risk to the target vehicle. We leverage (\ref{eq:dualUpdate_tv}) and (\ref{eq:dualUpdate_sv}) for the target vehicle and its surrounding connected vehicles, respectively. 
\begin{subequations}\label{eq:dualUpdate_tv}
    \begin{align}
\label{subeq:imp1}
\bm{p}_{[2, i]}^{i, k+1} & =\bm{p}_{[2, i]}^{i, k}+\rho(N-1)\left(\bm{y}_{[2, i]}^{i, k}-\bm{y}_{[2, i]}^k\right), \\ 
\label{subeq:imp2}
\bm{s}_{[2, i]}^{i, k+1} & =\bm{s}_{[2, i]}^{i, k}+\sigma\left(\bm{y}_{[2, i]}^{i, k}-\bm{x}_{[2, i]}^{i, k}\right), \\ 
\bm{r}_{[2, i]}^{i, k+1} & =\sigma \bm{x}_{[2, i]}^{i, k}+\rho(N-1)\left(\bm{y}_{[2, i]}^{i, k}+\bm{y}_{[2, i]}^k\right)\notag\\
\label{subeq:imp3}
&-(\bm{k}_{[2, i]}^{i, k+1}+\bm{p}_{[2, i]}^{i, k+1}+\bm{s}_{[2, i]}^{i, k+1}), \\
\label{subeq:imp4}
\bm{y}_{[2, i]}^{i, k+1} & =2\gamma\left(\bm O^i \bm{\Delta Z}^{i, k+1}+\bm{r}_{[2, i]}^{i, k+1}\right), \\
\label{subeq:imp5}
\bm{x}_{[2, i]}^{i, k+1} & =\Pi_{\mathcal{S}^{b\circ}}\left(\frac{1}{\sigma}\left(\bm{s}_{[2, i]}^{i, k+1}+\bm{y}_{[2, i]}^{i, k+1}\right)\right) .
\end{align}
\end{subequations}
Note that for the dual update of $\bm{y}_{[2, i]}^{v, k+1}$ of the surrounding vehicles connected to the target vehicle, they only need to consider the consensus to $\bm{r}_{[2, i]}^{k+1}$ in (\ref{eq:sub4}). For comparison, the target vehicle needs to consider the box constraints by calculating the high-dimensional matrix multiplication in (\ref{subeq:imp4}), which is time-consuming and meaningless.
\begin{subequations}\label{eq:dualUpdate_sv}
    \begin{align}
\label{eq:sub1}
\bm{p}_{[2, i]}^{v, k+1} & =\bm{p}_{[2, i]}^k+\rho\left(\bm{y}_{[2, i]}^k-\bm{y}_{[2, i]}^{i, k}\right) = \bm{p}_{[2, i]}^{k+1}, \\
\label{eq:sub2}
\bm{s}_{[2, i]}^{v, k+1} & =\bm{s}_{[2,1]}^k+\sigma\left(\bm{y}_{[2, i]}^k-\bm{x}_{[2, i]}^k\right)=\bm{s}_{[2, i]}^{k+1}, \\
\bm{r}_{[2, i]}^{v, k+1} & =\sigma \bm{x}_{[2, i]}^k+\rho(2 |n^v|-1) \bm{y}_{[2, i]}^k+\rho \bm{y}_{[2, i]}^{i, k}\notag  \\
&\label{eq:sub3}
-(\bm{k}_{[2, i]}^{k+1}+\bm{p}_{[2, i]}^{k+1}+\bm{s}_{[2, i]}^{k+1})=\bm{r}_{[2, i]}^{k+1},\\
\label{eq:sub4}
\bm{y}_{[2, i]}^{v, k+1} & =2\gamma \bm{r}_{[2, i]}^{k+1}=\bm{y}_{[2, i]}^{k+1}, \\
\label{eq:sub5}
\bm{x}_{[2, i]}^{v, k+1} & =\Pi_{\mathcal{S}^{b\circ}}\left(\frac{1}{\sigma}\left(\bm{s}_{[2, i]}^{k+1}+\bm{y}_{[2, i]}^{k+1}\right)\right)=\bm{x}_{[2, i]}^{k+1},
\end{align}
\end{subequations}

For the primal update process, due to the fully decentralized feature of the algorithm, only one target vehicle is considered in a standard LQR problem formulated as:
\begin{equation}
\begin{aligned}
    \min_{\bm {\Delta Z^i}}\, &\bm{\Delta Z}^{i\top}\bm L^i_1+\frac{1}{2}\bm{\Delta Z}^{i\top}\bm L^i_2\bm{\Delta Z}^i + 2\bm r^{i,k+1\top}\bm J^i\bm{\Delta Z}^{i}\\
    &+\frac{1}{\sigma+2\rho |n^i|}\bm{\Delta Z}^{i\top}\bm J^{i\top}\bm J^i\bm{\Delta Z}^{i}  \\
    \text{s.t. }\, &\left(\bm L^i_3-\bm L^i_4\right) \bm{\Delta Z}^i=0.
\end{aligned}
\label{eq:quadOptProb}
\end{equation}

Furthermore, as depicted in the Algorithm~\ref{alg:alg3}, in the scope of task execution, for the vehicles in the AMoD system, the vehicles need to evolve the connection graph guided by the LMM and solve the cooperative motion planning problems parallelly using the receding horizon manner.

% It is noted that the LMM-guided graph evolution is executed on a periodic basis. Should this process be integrated into a real-world or real-time performance setting, the states of the CAVs can be predicted for a future time horizon $T_e$, and the corresponding BEV data can be transmitted to the LMM. During the waiting period for the LMM's response, the pre-planned motion can be executed concurrently.
\section{Simulation Results}
\begin{figure*}[htbp]
\centering
\subfigure[The Initialized BEV of the AMoD]{
\includegraphics[ width=0.4\linewidth]{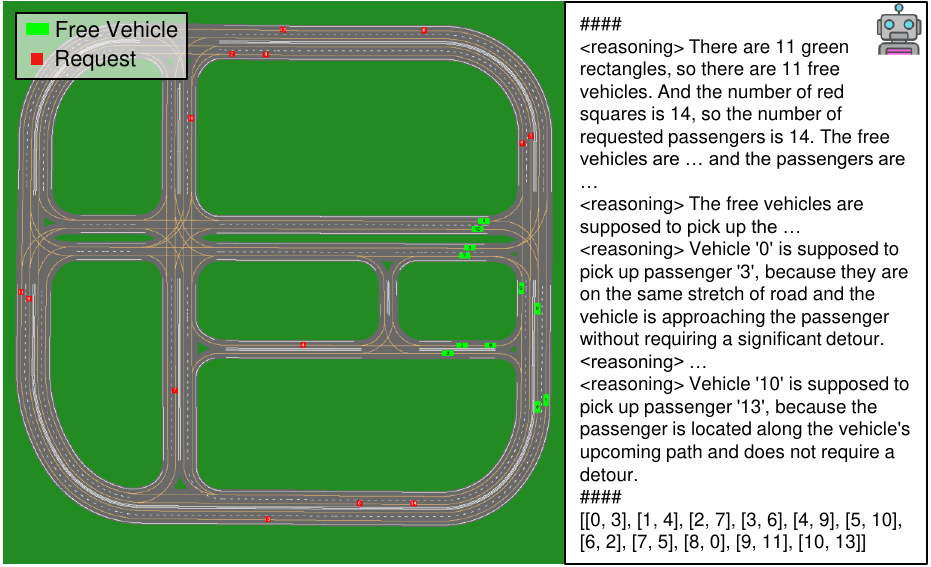}
\label{subfig:step0}
}
\subfigure[The BEV of the AMoD after 20 steps]{
\includegraphics[width=0.4\linewidth]{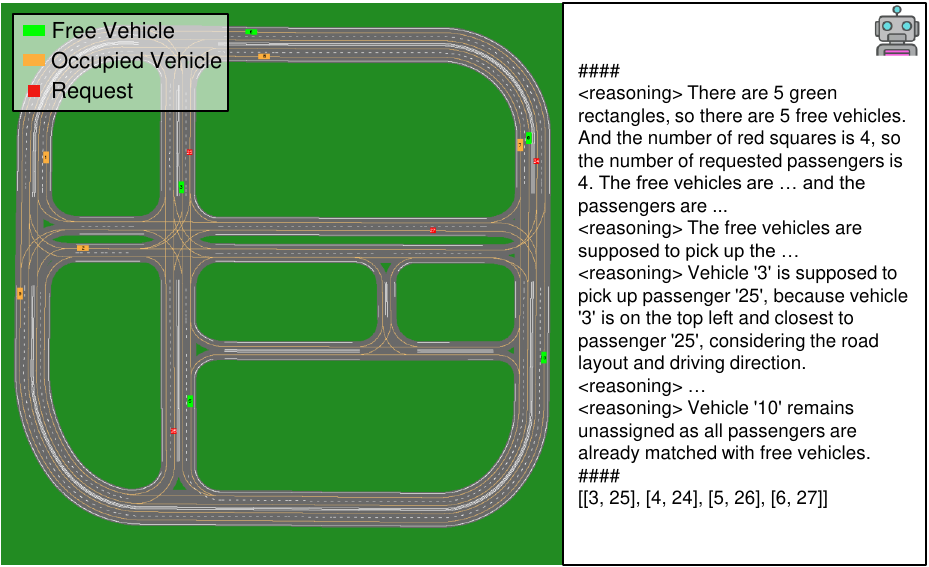}
\label{subfig:step20}
}
% \DeclareGraphicsExtensions.
\caption{The reasoning demonstration of the AMoD system utilizing our proposed LMMCoDrive framework: Green rectangles represent free vehicles, whereas yellow rectangles indicate vehicles occupied by passengers. Red squares signify pending requests from passengers within the AMoD system. The corresponding responses from the LMM are also revealed alongside the BEVs.}
\label{fig:ScheduleSimulation}
\end{figure*}
% Table generated by Excel2LaTeX from sheet 'Sheet1'
\begin{table*}[htbp]
  \centering
  \caption{Comparison of scheduling efficiency between LMMCoDrive and two heuristic algorithms}
    \begin{tabular}{ccccccccccccc}
    \toprule
    \multirow{2}[4]{*}{Method} & \multicolumn{3}{c}{Episode 1} & \multicolumn{3}{c}{Episode 2} & \multicolumn{3}{c}{Episode 3} & \multicolumn{3}{c}{Episode 4} \\
\cmidrule{2-13}          & $T_{atc}$ & $T_{stc}$ & $T_{mtc}$ & $T_{atc}$ & $T_{stc}$ & $T_{mtc}$ & $T_{atc}$ & $T_{stc}$ & $T_{mtc}$ & $T_{atc}$ & $T_{stc}$ & $T_{mtc}$ \\
    \midrule
    FCFS  & 16.06 & 5.73  & 22.78 & 11.15 & 5.43  & 14.64 & 17.74 & 3.15  & 22.98 & 9.93  & 8.30   & 18.24 \\
    DF    & 10.72 & 6.69  & 22.97 & 4.81  & 5.66  & 14.61 & 5.83  & 4.05  & 12.31 & 9.93  & 8.30   & 18.24 \\
    LMMCoDrive & 13.23 & 5.72  & 19.78 & 3.49  & 3.95  & 10.34 & 13.46 & 3.17  & 18.23 & 8.42  & 7.06  & 15.48 \\
    \bottomrule
    \end{tabular}%
  \label{tab:addlabel}%
\end{table*}%

\subsection{Environment and Evaluation Metrics}
We comprehensively evaluate the proposed LMMCoDrive framework in an AMoD system. For the scheduling part of the pairs of passengers and vehicles, we compare our proposed method with several popular heuristic algorithms, First Come First Serve (FCFS) and Distance First (DF). In terms of the graph evolution part for the vehicles, we execute an ablation study of our proposed method with the conditional Manhattan Distance-based method~\cite{liu2024improved}. The out-of-box LMM is GPT-4o. All the experiments are executed in $\texttt{Town10HD}$ of CARLA~\cite{dosovitskiy2017carla} with the assistance of LangChain in Python.
% We construct the AMoD system based on the Town10 provided by CARLA. The main objects of the AMoD are pedestrians making requests, free vehicles to be assigned by the scheduler, and occupied vehicles by passengers. The main objectives of the system are to convey passengers to their destinations safely and efficiently, and respond to requests from pedestrians as soon as possible.
% The evaluation metrics are divided into two parts, scheduling and cooperative driving. For scheduling performance, we use task completion time $T_{tc}$ to assess the overall performance of LMMCoDrive. Note that we calculate the average time, standard deviation, and maximum time, which are denoted as $T_{atc}$ and $T_{mtc}$.
The evaluation metrics are divided into two categories: scheduling and cooperative driving. For scheduling performance, we utilize the task completion time $ T_{tc} $ to assess the overall effectiveness of LMMCoDrive. Specifically, we calculate the average task completion time $T_{atc}$, the standard deviation $T_{stc}$, and the maximum task completion time $T_{mtc}$. The computational time for each episode of cooperative driving is utilized to represent the effectiveness of the LMM-guided graph evolution strategy.
\subsection{Scheduling Performance}\label{subsec:schedulingPerformance}

We conduct the comparisons in the aforementioned AMoD system with 10 CAVs, and requests from passengers from all over the traffic system are gradually raised. 
% As shown in Fig.~\ref{fig:ScheduleSimulation}, the CAVs denoted with rectangles pick up the passengers at their original position. 
As shown in Fig.~\ref{fig:ScheduleSimulation}, the vehicles convey passengers to their randomly generated destinations with the yellow remark. As for the free vehicles, once they are assigned a passenger in the AMoD system, they will drive to their customs and serve them as soon as possible.
As depicted in Table~\ref{tab:addlabel}, in most cases, LMMCoDrive has the smallest $T_{mtc}$ and $T_{stc}$, while maintaining good $T_{atc}$. It proves the excellent scheduling performance for the AMoD. 
% The LMMCoDrive achieves the balance between efficiency and passenger experience. Nevertheless, in some cases, LMMCoDrive \hl{performs unstable denoting} that the prompt of the LMM can be further refined.
The LMMCoDrive achieves a balance between efficiency and passenger experience, but performs unstable in some cases, and this indicates that the prompt of the LMM can be further refined.

\subsection{Ablation Study}
\begin{figure}
    \centering
    \includegraphics[width=1\linewidth]{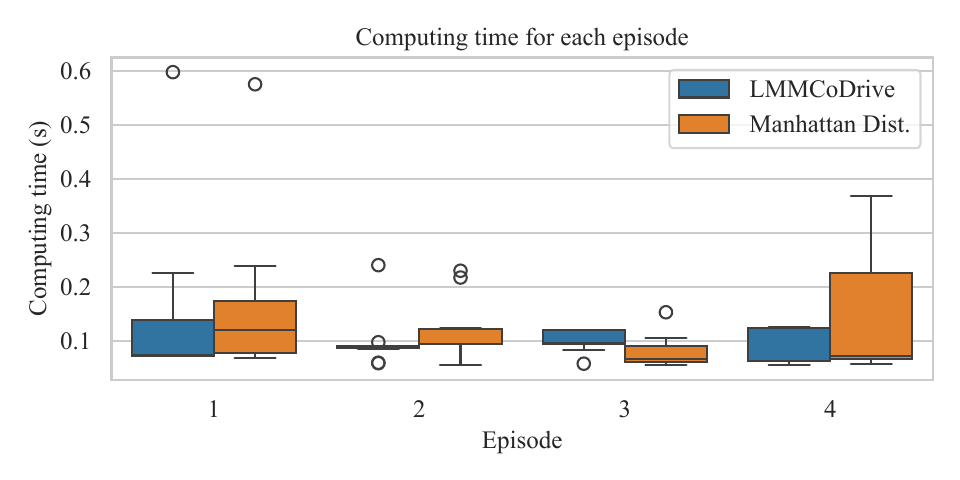}
    \caption{The computation time for each episode of the receding horizon for cooperative motion planning. For the first column of boxes, the graph of the connection of the vehicles is evolved by LMM, while the second column is the result of the Manhattan distance-based heuristic method.}
    \label{fig:computingGraphEvo}
\end{figure}

% The purpose of this ablation study is to estimate the effectiveness of the LMM for graph evolution for the AMoD systems. As the default graph evolution setting, a heuristic strategy similar to methods in Section~\ref{subsec:schedulingPerformance}. As shown in Fig.~\ref{fig:computingGraphEvo}, because of the accurate and flexible decision of the groups of CAVs having collision risks, the average computation time of the parallel OCPs is obviously lower than the Manhattan distance-based graph-evolution method. The main reason is that the LMMCoDrive can reason and reflect the potential collisions not too conservative.
The purpose of this ablation study is to evaluate the effectiveness of the LMM in the graph evolution process for AMoD systems. By default, the graph evolution setting employs a heuristic strategy similar to those detailed in Section~\ref{subsec:schedulingPerformance}. As illustrated in Fig.~\ref{fig:computingGraphEvo}, the LMM's capability to accurately and flexibly determine groups of CAVs with collision risks results in a significantly lower average computation time for parallel OCPs compared to the Manhattan distance-based graph-evolution method. The primary reason for this improvement is that LMMCoDrive can more precisely identify potential collisions, thereby avoiding overly conservative decisions that would otherwise increase the number of CAVs in one OCP leading to computational overhead.

\section{Conclusion}
In this paper, we propose LMMCoDrive, an LMM-enabled integrated scheduling and cooperative motion planning framework suitable for AMoD systems in urban scenarios.
The incorporation of a BEV that describes the traffic situation expands the boundary of the applications of LLMs, and this alleviates the travel prompt engineering. 
% The ADMM-based cooperative driving algorithm is seamlessly aggregated.
Simulation results demonstrate LMM's role in optimizing CAV scheduling and facilitating decentralized cooperative optimization. It's a significant milestone towards practical and efficient AMoD systems, which also offers insights for future urban transportation solutions. Future work will focus on further refining its information flow to achieve a more robust performance and expanding its application to larger and more complex urban cooperative driving scenarios.

% \addtolength{\textheight}{-12cm}   % This command serves to balance the column lengths
%                                   % on the last page of the document manually. It shortens
%                                   % the textheight of the last page by a suitable amount.
%                                   % This command does not take effect until the next page
%                                   % so it should come on the page before the last. Make
%                                   % sure that you do not shorten the textheight too much.

%%%%%%%%%%%%%%%%%%%%%%%%%%%%%%%%%%%%%%%%%%%%%%%%%%%%%%%%%%%%%%%%%%%%%%%%%%%%%%%%

%%%%%%%%%%%%%%%%%%%%%%%%%%%%%%%%%%%%%%%%%%%%%%%%%%%%%%%%%%%%%%%%%%%%%%%%%%%%%%%%

%%%%%%%%%%%%%%%%%%%%%%%%%%%%%%%%%%%%%%%%%%%%%%%%%%%%%%%%%%%%%%%%%%%%%%%%%%%%%%%%

%%%%%%%%%%%%%%%%%%%%%%%%%%%%%%%%%%%%%%%%%%%%%%%%%%%%%%%%%%%%%%%%%%%%%%%%%%%%%%%%

\bibliographystyle{IEEEtran}
\bibliography{refs}

\end{document}